\documentclass[runningheads]{llncs}
\usepackage{graphicx}
\usepackage{comment}
\usepackage{amsmath,amssymb} % define this before the line numbering.
\usepackage{color}
\usepackage{multirow}
\usepackage{subcaption}

\usepackage{url}

\usepackage[pagebackref=true,breaklinks=true,letterpaper=true,colorlinks,bookmarks=false]{hyperref}

\usepackage{xspace}
\makeatletter
\DeclareRobustCommand\onedot{\futurelet\@let@token\@onedot}
\def\@onedot{\ifx\@let@token.\else.\null\fi\xspace}

\def\eg{\emph{e.g}\onedot}

 \def\vs{\emph{vs}\onedot}
 
\def\etal{\emph{et al}\onedot}
\makeatother

\begin{document}
% \linenumbers
\pagestyle{headings}
\mainmatter
\def\ECCVSubNumber{4374}  % Insert your submission number here

% \title{Eff-SpineNet: Building SpineNet with Efficient Building s (placeholder)} % Replace with your title
\title{Efficient Scale-Permuted Backbone with \\ Learned Resource Distribution}

%\begin{comment}
%\titlerunning{ECCV-20 submission ID \ECCVSubNumber} 
%\authorrunning{ECCV-20 submission ID \ECCVSubNumber} 
%\author{Anonymous ECCV submission}
%\institute{Paper ID \ECCVSubNumber}
%\end{comment}
%******************

%\begin{comment}
\titlerunning{Efficient Scale-Permuted Backbone with Learned Resource Distribution}
\author{
Xianzhi Du \and Tsung-Yi Lin\and
Pengchong Jin\and
Yin Cui\\
Mingxing Tan\and
Quoc Le\and
Xiaodan Song
}
\authorrunning{X. Du et al.}
\institute{Google Research, Brain Team\\ \email{\{xianzhi,tsungyi,pengchong,yincui,tanmingxing,qvl,xiaodansong\}@google.com}}


%\end{comment}
%******************
\maketitle

\begin{abstract}
Recently, SpineNet has demonstrated promising results on object detection and image classification over ResNet model. However, it is unclear if the improvement adds up when combining scale-permuted backbone with advanced efficient operations and compound scaling. Furthermore, SpineNet is built with a uniform resource distribution over operations. While this strategy seems to be prevalent for scale-decreased models, it may not be an optimal design for scale-permuted models. In this work, we propose a simple technique to combine efficient operations and compound scaling with a previously learned scale-permuted architecture. We demonstrate the efficiency of scale-permuted model can be further improved by learning a resource distribution over the entire network. The resulting efficient scale-permuted models outperform state-of-the-art EfficientNet-based models on object detection and achieve competitive performance on image classification and semantic segmentation. Code and models will be open-sourced soon.

%\xianzhi{1). Mention NAS in the abstract?
%2). There are two improvements in SpineNet's architecture: First, we use space-to-depth as the main down-sampling method instead of stride-2 convolution; Second, we implement weighted block fusion throughout the network. We can briefly mention the two points in the abstract.}

%\ty{(1) Maybe let's add NAS for learning distribution part. (2) I think this is important. But let's leave these two improvements in the method section. We should have a good figure to compare re-sampling method in v1 and v2.}

\keywords{Scale-Permuted Model, Object Detection}
\end{abstract}

%%%%%%%%%%%%%%%%%%%%%%%%%%%%%%%%%%%%%%%%%%%%%
\begin{figure}
\centering
\begin{subfigure}[t]{0.5\textwidth}
        \centering
        \includegraphics[height=1.65in]{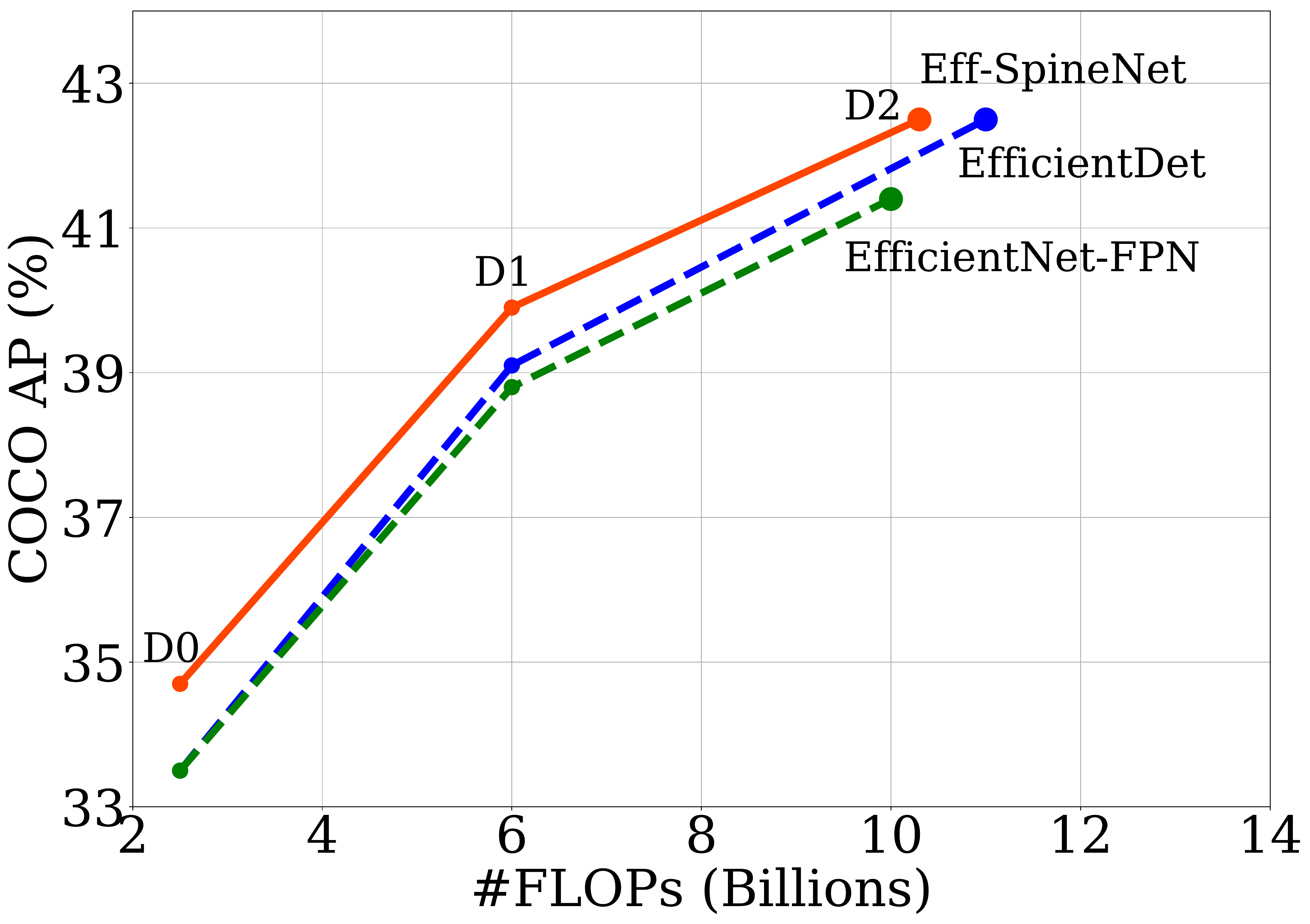}
        %\captionsetup{labelformat=empty}
        % \caption{Regular-size models}
    \end{subfigure}%
\begin{subfigure}[t]{0.5\textwidth}
        \centering
        \includegraphics[height=1.65in]{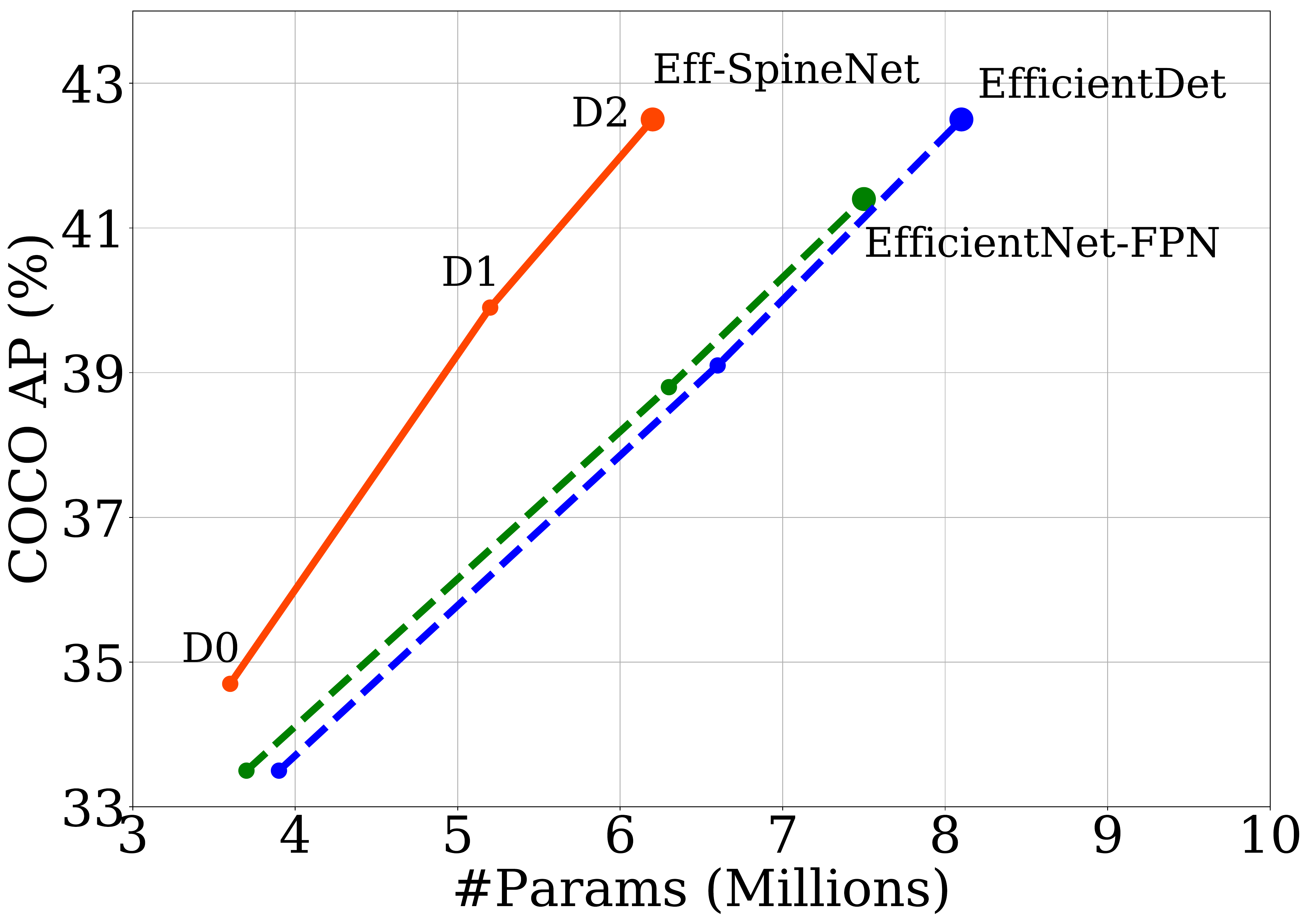}
        %\captionsetup{labelformat=empty}
        % \caption{Mobile-size models}
    \end{subfigure}%
\caption{Eff-SpineNet achieves better FLOPs \vs AP and Params \vs AP trade-off curves for regular-size object detection comparing to state-of-the-art scale-decreased models EfficientNet-FPN and EfficientDet. All models adopt the RetinaNet framework~\cite{retinanet}}
\label{fig:flops_ap_curve}
\end{figure}
%%%%%%%%%%%%%%%%%%%%%%%%%%%%%%%%%%%%%%%%%%%%%

\section{Introduction}

The scale-permuted network proposed by Du \etal~\cite{Du2020SpineNet} opens up the design of a new family of meta-architecture that allows wiring features with a scale-permuted ordering in convolutional neural network. The scale-permuted architecture achieves promising results on visual recognition and localization by significantly outperforming its scale-decreased counterpart when using the same residual operations but different architecture topology. Concurrently, EfficientNet-based models~\cite{tan2019efficientnet,tan2020efficientdet} demonstrate state-of-the-art performance using an advanced MBconv operation and the compound model scaling rule, while still adopting a scale-decreased backbone architecture design. A natural question is: \textit{can we obtain new state-of-the-art performance by combining scale-permuted architecture and efficient operations?}

In this paper, we decompose the model design into three parts: (1) architecture topology; (2) operation; (3) resource distribution. The architecture topology describes the wiring and the resolution of features. The operation defines the transformation (\eg, convolution and ReLU) applied to the features. The resource distribution indicates the computation allocated for each operation. Our study begins with directly combining the scale-permuted architecture topology from~\cite{Du2020SpineNet} and efficient operations from~\cite{tan2019efficientnet}. Unlike the previous works, we purposely \textit{do not} perform any neural architecture search because the architecture topology and operation have been extensively studied and learned by sophisticated neural architecture search algorithms respectively. Instead of designing a joint search space for learning an even more tailored model, we are curious if the scale-permuted architecture and efficient operations are \textit{generic} in the status quo and can directly be used to build the state-of-the-art model.

Despite having the learned advanced architecture topology and operation, the resource distribution has not been well studied in isolation in existing works. In~\cite{Du2020SpineNet}, the resource distribution is nearly uniform for all operations, regardless of the resolution and location of a feature in the architecture. In~\cite{tan2019efficientnet}, the search space contains only a few hand-selected feature dimensions for each operation and the neural architecture search algorithm is learned to select the best one. This greatly limits the possible resource distribution over the entire network. In this work, we propose a search algorithm that learns the resource distribution with the fixed architecture topology and operation. Given the target resource budget, we propose to learn the percentage of total computation allocated to each operation. In contrast to learning the absolute feature dimension, our resource targeted algorithm has the advantage of exploring a wider range of resource distribution in a manageable search space size.

We mainly conduct experiments on object detection using COCO dataset~\cite{coco}. We carefully study the improvements brought by the architecture topology and operation and discover that simply combining scale-permuted architecture and MBConv operation outperforms EfficientDet~\cite{tan2020efficientdet}. The experiment results show that the architecture topology and operation are complementary for improving performance. We show that the scale-permuted EfficientNet backbone, which shares the same operation but different architecture topology with EfficientNet-FPN, improves the performance across various models and input image sizes while using less parameters and FLOPs. We further improve the performance by learning a resource distribution for the scale-permuted EfficientNet backbone. The final model is named Efficient SpineNet (Eff-SpineNet). We discover that the model prefers to distribute resources unevenly to each operation. Surprisingly, the best resource distribution saves 18\% of model parameters given the similar FLOPs, allowing us to build a more compact model.

Lastly, we take Eff-SpineNet and evaluate its performance on image classification and semantic segmentation. Eff-SpineNet achieves competitive results on both tasks. 
Interestingly, we find that Eff-SpineNet is able to retain the performance with less parameters. 
Compared with EfficientNet that is specifically designed for image classification, Eff-SpineNet has around 35\% less parameters under the same FLOPs, while the Top-1 ImageNet accuracy drops by less than 1-2\%.
For semantic segmentation, Eff-SpineNet models achieve comparable mIOU on PASCAL VOC val 2012 to popular semantic segmentation networks, such as the DeepLab family~\cite{Chen2017deeplabv3,Chen_2018_deeplabv3plus}, while using 95\% less FLOPs.
To summarize, these observations show that Eff-SpineNet is versatile and is able to transfer to other visual tasks including image classification and semantic segmentation.
% This observation shows that the behavior of scale-permuted and scale-decreased model are quite different.
% \xianzhi{Not very confident about this claim. All our search jobs are biased to object detection. The less-params design may not be good for classification.}
% The scale-permuted model architecture may offer the new trade-off between FLOPs, parameters, and accuracy for the future research.

\section{Related Work}
\paragraph{\bf Scale-permuted network:} Multi-scale feature representations have been the core research topic for object detection and segmentation. The dominant paradigm is to have a strong backbone model with a lightweight decoder such as feature pyramid networks~\cite{fpn}. Recently, many work has discovered performance improved with a stronger decoder~\cite{nasfpn,liu2018panet,zhao2019m2det}. Inspired by NAS-FPN~\cite{nasfpn}, SpineNet~\cite{Du2020SpineNet} proposes a scale-permuted backbone architecture that removes the distinction of encoder and decoder and allows the
scales of intermediate feature maps to increase or decrease anytime, and demonstrates promising performance on object detection and image classification. Auto-DeepLab~\cite{autodeeplab} is another example that builds scale-permuted models for semantic segmentation.

\paragraph{\bf Efficient operation:} The efficiency is the utmost important problem for mobile-size convolution model. The efficient operations have been extensively studied in the MobileNet family~\cite{mobilenetv2,mobilenetv3,mobilenetv3,mnasnet}. Spare depthwise convolution and the inverted bottleneck block are the core ideas for efficient mobile size network. MnasNet~\cite{mnasnet} and EfficientNet~\cite{tan2019efficientnet} takes a step further to develop MBConv operation based on the mobile inverted bottleneck in~\cite{mobilenetv2}. EfficientNet shows that the models with MBConv operations not only achieving the state-of-the-art in ImageNet challenge but also very efficient. Recently, EfficientDet~\cite{tan2020efficientdet} builds object detection models based on the EfficientNet backbone model and achieves impressive detection accuracy and computation efficiency.

\paragraph{\bf Resource-aware neural architecture search:} In neural architecture search, adding resource constraints is critical to avoid the bias to choose a model with higher computation. MnasNet~\cite{mnasnet} introduces multi-objective rewards that optimize the model accuracy while penalizes models that violate the constraints. CR-NAS~\cite{Liang2020Computation} searches for the best resource allocation by learning the number of blocks allocated in each resolution stage and the dilated convolution kernel.

\section{Method}

In this section, we first describe how to combine the scale-permuted architecture topology~\cite{Du2020SpineNet} and efficient operation MBConv~\cite{tan2019efficientnet}. Then, we introduce feature resampling and fusion operations in the efficient scale-permuted model. Lastly, we propose a search method to learn resource distribution for building Eff-SpineNet.

\subsection{Scale-permuted Architecture with Efficient Operations}
We first combine SpinetNet-49 architecture topology with MBconv blocks. We start with permuting the EfficientNet-B0 model. The goal here is to build an efficient scale-permuted model, SP-EfficientNet-B0, that has the similar computation and parameters as the EfficientNet-B0 baseline. We follow the idea of the compound scaling rule in EfficientNet to create 5 higher capacity models.

\begin{figure}[t]
\centering
    \includegraphics[width=0.8\textwidth]{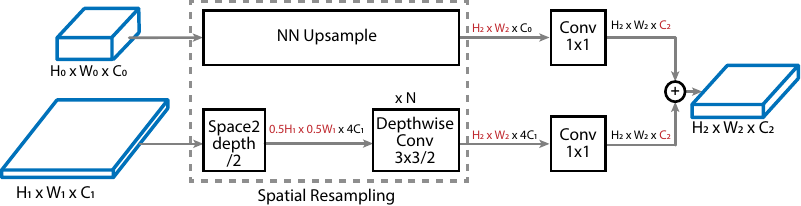}
    \caption{Resampling operation}
\label{fig:resampling}
\end{figure}

\paragraph{\bf SP-EfficientNet-B0:} We attempt to replace all the residual and bottleneck blocks in SpineNet-49 with MBconv blocks. One design decision is how to assign the convolution kernel size and feature dimension when applying MBConv to scale-permuted architecture. Given SpineNet-49 has already had a large receptive field, we decide to fix the kernel size to 3 for all MBConv operations. To obtain a model with similar size as EfficientNet-B0, we obtain the feature dimension for each level by averaging the feature dimensions over all blocks at the corresponding levels in Efficient-B0. Since the $L_6$ and $L_7$ blocks does not have a corresponding feature in EfficientNet, we follow~\cite{Du2020SpineNet} to set them to have the same feature dimension as the $L_5$ block. The detailed network specifications of the SP-EfficientNet-B0 model is presented in Table~\ref{table:architecture}. 

\paragraph{\bf Compound scaling for scale-permuted network:} We follow the compound scaling rule proposed in~\cite{tan2019efficientnet} to scale up the SP-EfficientNet-B0 model. We use the rule to compute the number of blocks for each feature level, feature dimension, and input image size. Since the number of blocks for a level after scaling may be more than the blocks at the corresponding level in SP-EfficientNet-B0 model, we uniformly repeat the blocks in SP-EfficientNet-B0 model. If the scaled number of blocks is not the multiple of those in SP-EfficientNet-B0, we add the remainder blocks one-by-one in the bottom up ordering. The detailed model scaling specifications are given in Table~\ref{table:scaling_map}.

%%%%%%%%%%%%%%%%%%%%%%%%%%%%%%%%%%%%%%%%%%%%%
\setlength{\tabcolsep}{4pt}
\begin{table}[t]
\begin{center}
\caption{Block specifications for EfficentNet-B0, SP-EfficientNet-B0, and Eff-SpineNet-D0, including block level, kernel size, and output feature dimension. SP-EfficientNet-B0 and Eff-SpineNet-D0 share same specifications for block level and kernel size}
\label{table:architecture}
\scalebox{0.85}{
\begin{tabular}{c | ccc | cccc}
\hline
\multirow{3}{*}{block id} & \multicolumn{3}{c|}{EfficientNet-B0} & \multicolumn{4}{c}{scale-permuted models} \\
 \cline{2-8}
 &  \multirow{2}{*}{level}  & \multirow{2}{*}{kernel} & \multirow{2}{*}{feat. dim} & \multirow{2}{*}{level}  & \multirow{2}{*}{kernel} & \multicolumn{2}{c}{feat. dim}\\
 & & & & & & SP-EfficientNet-B0 & Eff-SpineNet-D0 \\
 \cline{2-7}
 \hline
 1  & $L_1$ & $3\times3$ & 16 & $L_1$ & $3\times3$ & 16 & 16 \\
 2  & $L_2$ & $3\times3$ & 24 & $L_2$ & $3\times3$ & 24 & 24 \\
 3  & $L_2$ & $3\times3$ & 24 & $L_2$ & $3\times3$ & 24 & 16 \\
 4  & $L_3$ & $5\times5$ & 40 & $L_2$ & $3\times3$ & 24 & 16 \\
 5  & $L_3$ & $5\times5$ & 40 & $L_4$ & $3\times3$ & 96 & 104 \\
 6  & $L_4$ & $3\times3$ & 80 & $L_3$ & $3\times3$ & 40 & 48 \\
 7  & $L_4$ & $3\times3$ & 80 & $L_4$ & $3\times3$ & 96 & 120 \\
 8  & $L_4$ & $3\times3$ & 80 & $L_6$ & $3\times3$ & 152 & 40 \\
 9  & $L_4$ & $5\times5$ & 112 & $L_4$ & $3\times3$ & 96 & 120 \\
 10& $L_4$ & $5\times5$ & 112 & $L_5$ & $3\times3$ & 152 & 168 \\
 11& $L_4$ & $5\times5$ & 112 & $L_7$ & $3\times3$ & 152 & 96 \\
 12& $L_5$ & $5\times5$ & 192 & $L_5$ & $3\times3$ & 152 & 192 \\
 13& $L_5$ & $5\times5$ & 192 & $L_5$ & $3\times3$ & 152 & 136 \\
 14& $L_5$ & $5\times5$ & 192 & $L_4$ & $3\times3$ & 96 & 104 \\
 15& $L_5$ & $5\times5$ & 192 & $L_3$ & $3\times3$ & 40 & 40 \\
 16& $L_5$ & $3\times3$ & 320 & $L_5$ & $3\times3$ & 152 & 136 \\
 17  & - & - & - & $L_7$ & $3\times3$ & 152 & 136 \\
 18  & - & - & - & $L_6$ & $3\times3$ & 152 & 40 \\
\hline
\end{tabular}
}
\end{center}
\end{table}
\setlength{\tabcolsep}{1.4pt}
%%%%%%%%%%%%%%%%%%%%%%%%%%%%%%%%%%%%%%%%%%%%%

%%%%%%%%%%%%%%%%%%%%%%%%%%%%%%%%%%%%%%%%%%%%%
\setlength{\tabcolsep}{4pt}
\begin{table}
\begin{center}
\caption{Model scaling method for Eff-SpineNet models. \textbf{input size:} Input resolution. \textbf{feat. mult.:} Feature dimension multiplier for convolutional layers in backbone. \textbf{block repeat:} Number of repeats for each block in backbone. The 18 blocks are ordered from left to right. \textbf{feat. dim.:} Feature dimension for separable convolutional layers in subnets. \textbf{\#layers:} Number of separable convolutional layer in subnets}
\label{table:scaling_map}
\scalebox{0.85}{
\begin{tabular}{c | c c c | c c}
\hline
 \multirow{2}{*}{model id} & \multicolumn{3}{c|}{scale-permuted backbone} & \multicolumn{2}{c}{subnets} \\
 \cline{2-6}
  & input size & feat. mult. & block repeat & feat. dim. & \#layers \\
 \hline
 M0 & 256 & 0.4 & \{1,1,1,1,1,1,1,1,1,1,1,1,1,1,1,1,1,1\} & 24 & 3\\
 M1 & 384 & 0.5 & \{1,1,1,1,1,1,1,1,1,1,1,1,1,1,1,1,1,1\} & 40 & 3\\
 M2 & 384 & 0.8 & \{1,1,1,1,1,1,1,1,1,1,1,1,1,1,1,1,1,1\} & 64 & 3\\
 \hline
 D0 & 512 & 1.0 & \{1,1,1,1,1,1,1,1,1,1,1,1,1,1,1,1,1,1\} & 64 & 3\\
 D1 & 640 & 1.0 & \{2,2,1,1,2,2,2,1,1,2,1,2,1,1,1,1,1,1\} & 88 & 3\\
 D2 & 768 & 1.1 & \{2,2,1,1,2,2,2,1,1,2,1,2,1,1,1,1,1,1\} & 112 & 3\\
 
\hline
\end{tabular}
}
\end{center}
\end{table}
\setlength{\tabcolsep}{1.4pt}
%%%%%%%%%%%%%%%%%%%%%%%%%%%%%%%%%%%%%%%%%%%%%

\subsection{Feature Resampling and Fusion}
Given the MBConv output feature dimension is much lower compared to residual and bottleneck blocks, we redesign the feature resampling method. And we adopt the fusion method from EfficientDet~\cite{tan2020efficientdet}.

\paragraph{\bf Resampling method:} Since MBConv has a small output feature dimension, it removes the need of the scaling factor $\alpha$ in SpineNet to reduce the computation. Compared to SpineNet, the 1x1 convolution to reduce input feature dimension is removed. Besides, we find using the space-to-depth operation followed by stride 2 convolutions preserves more information than the original design, with a small increase of computation. The new resampling strategy is shown in Figure~\ref{fig:resampling}.

\paragraph{\bf Weighted block fusion:} As shown in~\cite{tan2020efficientdet}, input features at different resolutions or network building stages may contribute unequally during feature fusion. We apply the fast normalized fusion strategy introduced in~\cite{tan2020efficientdet} to block fusion in SpineNet. The method is shown in Equation~\ref{eq:fusion}:

\begin{equation}
B^{out}=\frac{\sum_i w_i\times B^{in}_i}{ \sum_j w_j + 0.001},
\label{eq:fusion}
\end{equation}
where $B^{in}$ and $B^{out}$ represent the input blocks and the output block respectively. $w$ is the weight to be learned for each input block.

\subsection{Learning Resource Distribution}

%%%%%%%%%%%%%%%%%%%%%%%%%%%%%%%%%%%%%%%%%%%%%
\begin{figure}
\centering
\begin{subfigure}[t]{0.5\textwidth}
        \centering
        \includegraphics[height=1.5in]{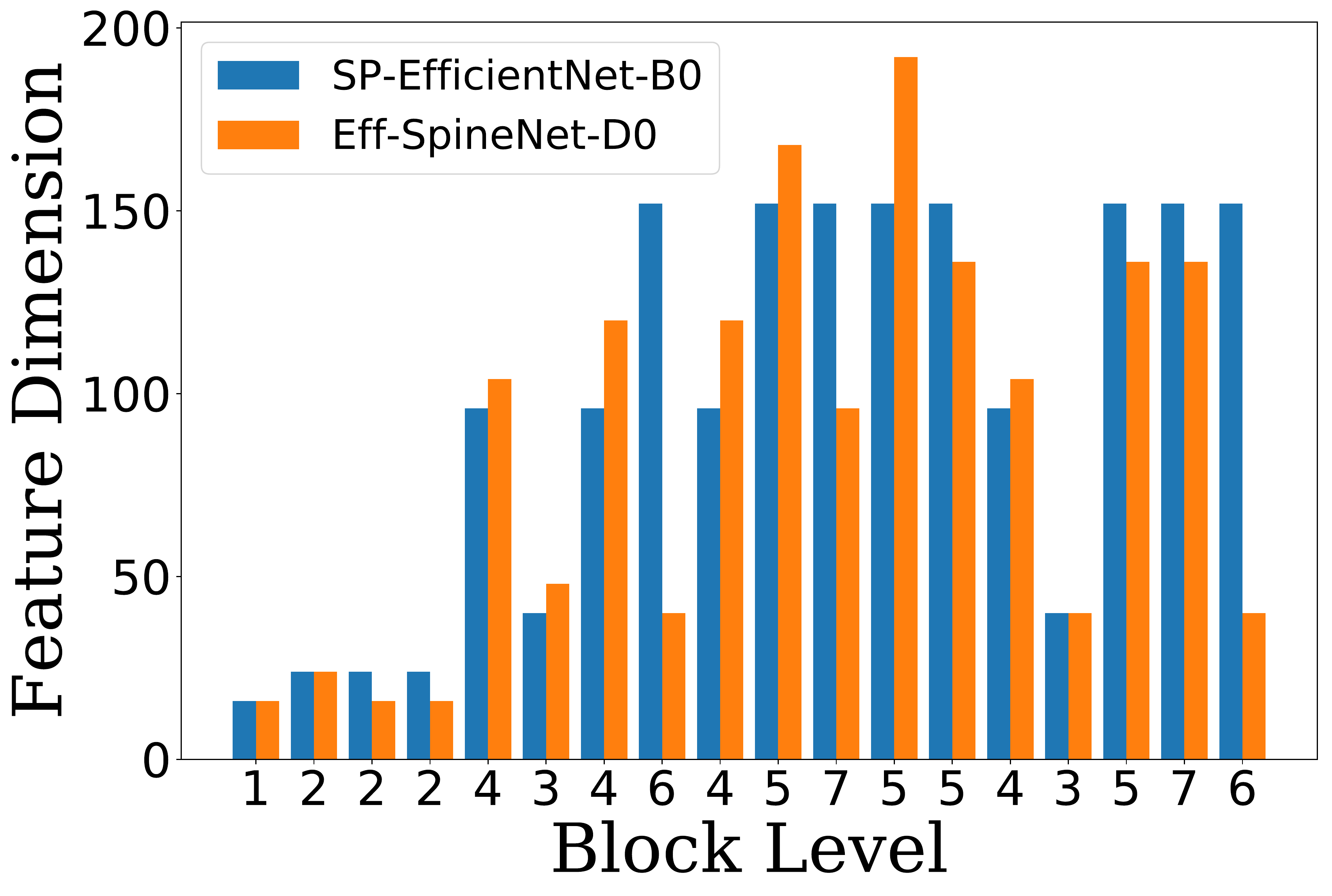}
        \captionsetup{labelformat=empty}
    \label{fig:feat_level}
    \end{subfigure}%
\begin{subfigure}[t]{0.5\textwidth}
        \centering
        \includegraphics[height=1.5in]{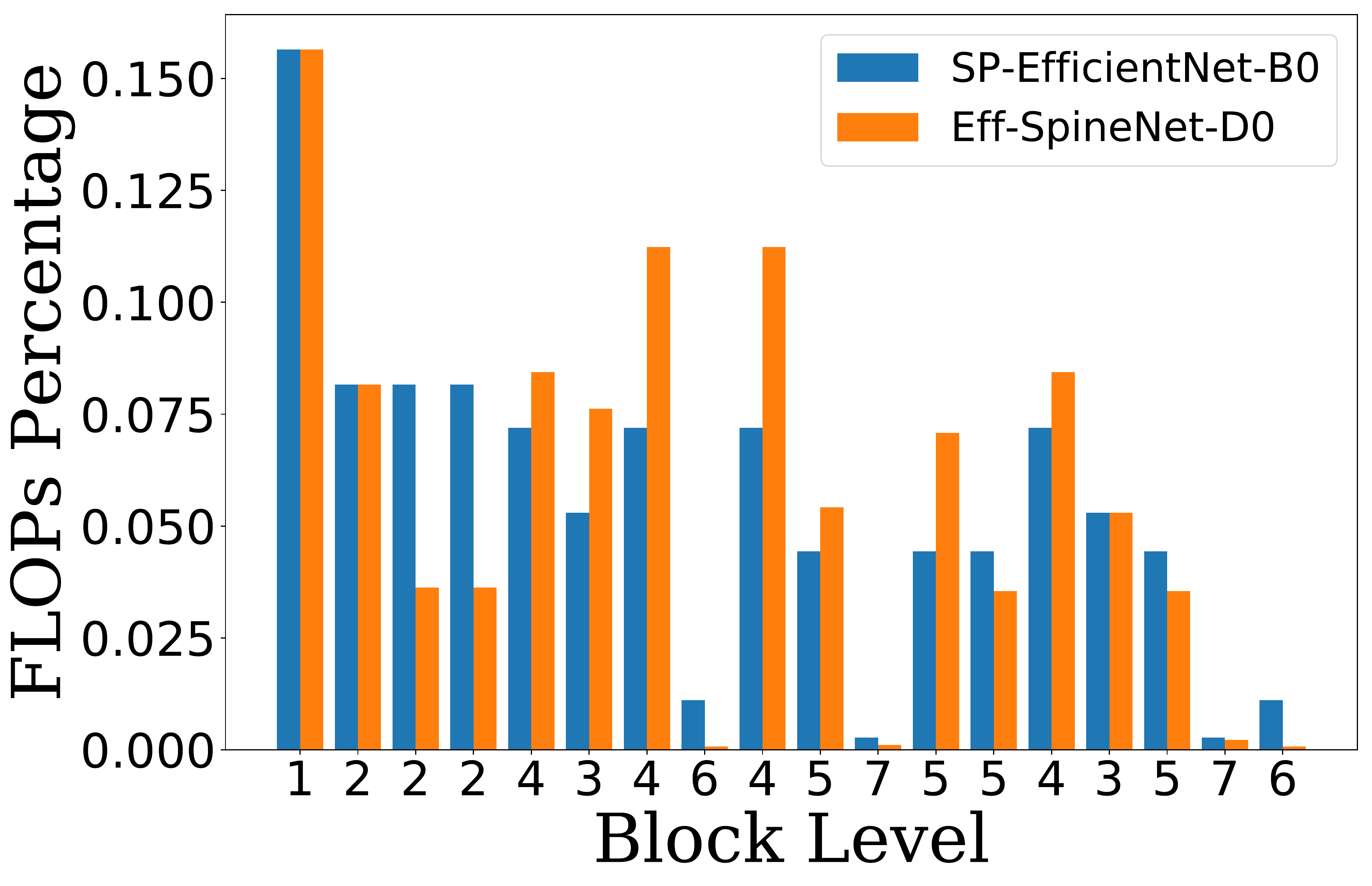}
        \captionsetup{labelformat=empty}
    \label{fig:comp_level}
    \end{subfigure}%
\caption{Comparisons of SP-EfficientNet-B0 and Eff-SpineNet-D0 in feature dimension distribution (left) and resource distribution (right). The 18 blocks are plotted in order from left to right with block level shown in the $x$-axis}
\label{fig:feat_level}
\end{figure}
%%%%%%%%%%%%%%%%%%%%%%%%%%%%%%%%%%%%%%%%%%%%%

Typically, the conventional architecture gradually increases feature dimension with decreasing spatial resolution of feature~\cite{resnet,resnext,tan2019efficientnet,mobilenetv2,mobilenetv3}. However, this design may be sub-optimal for a scale-permuted network. In this section, we propose a simple yet effective search method for resource reallocation. We learn the resource distribution through adjusting the feature dimension of MBConv blocks. In the search space, we fix the total FLOPs of MBConv blocks in the entire model, and learn a scale multiplier of feature dimension for each block in SP-EfficientNet-B0.

Consider $c_i$ to be the feature dimension of MBConv block $i$, the FLOPs can be computed as $\mathcal{F}_i\simeq C_i\times c_i^2$, where $C_i$ is a constant that depends on height, width, and expansion ratio of a given block.
\begin{equation}
\begin{aligned}
\mathcal{F}_i &= H_i\times W_i\times(2\times c_i^2\times r + k^2\times c_i\times r)\\
 &\simeq H_i\times W_i\times c_i^2\times 2\times r \\
 &\simeq C_i\times c_i^2,
\end{aligned}
\label{eq:flops}
\end{equation}
where $H_i$, $W_i$ is the height and width of the feature map, $r$ is the expansion ratio and $k$ is the kernel size in a MBConv block.

We propose to learn a multiplier $\alpha_i$ that adjusts the resource distribution over the entire model with a target total desired computation $\mathcal{F}_t$
\begin{equation}
\mathcal{F}_t = \sum_i \alpha_i \mathcal{F}_i
\label{eq:constraint}
\end{equation}
In our experiment, we simply set  $\mathcal{F}_t = \sum_i \mathcal{F}_i$.

Learning $\alpha_i$ can be challenging because $\alpha_i$ can be any positive real number. Here, we propose to learn a multiplier $\beta_i$ which is selected from a set of $N$ positive numbers $\{\beta^1, \beta^2, ..., \beta^N\}$. Then, we can represent $\alpha_i$ as a function of $\beta_i$ which satisfies the equation~\ref{eq:constraint}.

\begin{equation}
\alpha_i = \frac{\mathcal{F}_t}{\sum_k \beta_k \mathcal{F}_k} \beta_i
\end{equation}
Finally, we use $\alpha_i$ to modify the feature dimension for each block $\hat{c_i} = \sqrt{\alpha_i}c_i$.

Using this resource distribution learning strategy, we discover our final model, Eff-SpineNet-D0. We show the model specification in Table~\ref{table:architecture} and the comparison with  SP-EfficientNet-B0 in Figure~\ref{fig:feat_level}.

%%%%%%%%%%%%%%%%%%%%%%%%%%%%%%%%%%%%%%%%%%%%%
%\begin{figure}
%\centering
%  \includegraphics[height=6.5cm]{Eff-SpineNet/imgs/spinenet_img1.pdf}
%\caption{A placeholder. From EfficientNet to Eff-SpineNet. \ty{We don't need this. But we need a figure to explain %architecture toplogy, operation, and resource distribution.}}
%\label{fig:example}
%\end{figure}
%%%%%%%%%%%%%%%%%%%%%%%%%%%%%%%%%%%%%%%%%%%%%

\section{Applications}

\subsection{Object Detection}

We use Eff-SpineNet as backbone in RetinaNet~\cite{retinanet} for one-stage object detection and in Mask R-CNN~\cite{mrcnn} for two-stage object detection and instance segmentation. The  feature map of the 5 output blocks \{$P_3$, $P_4$, $P_5$, $P_6$, $P_7$\} are used as the multi-scale feature pyramid. Similar to~\cite{tan2020efficientdet}, we design a heuristic scaling rule to maintain a balance in computation between backbone and subnets during model scaling and use separable convolutions in all subnets. In RetinaNet, we gradually use more convolutional layers and a larger feature dimension for each layer in the box and class subnets for a larger Eff-SpineNet model. In Mask R-CNN, the same scaling rule is applied to convolutional layers in the RPN, Fast R-CNN and mask subnets. In addition, a fully connected layer is added after convolutional layers in the Fast R-CNN subnet and we apply the scaling to adjust its dimension to 256 for D0 and D1, and 512 for D2. Details are shown in Table~\ref{table:scaling_map}.

\subsection{Image Classification}\label{sec:image_classification}
We directly utilize all feature maps from $P_3$ to $P_7$ to build the classification model.
Different from the object detection models shown in Table~\ref{table:scaling_map}, we set the feature dimension to $256$ for all models.
The final feature vector is generated by nearest-neighbor upsampling and averaging all feature maps to the same size as $P_3$ followed by the global average pooling.
We apply a linear classifier on the $256$-dimensional feature vector and train the classification model with softmax cross-entropy loss.

\subsection{Semantic Image Segmentation}\label{sec:semantic_seg}
In this subsection, we explore Eff-SpineNet for the task of semantic image segmentation. We apply nearest-neighbor upsampling to match the sizes of all feature maps in \{$P_3$, $P_4$, $P_5$, $P_6$, $P_7$\} to $P3$ then take the average. The averaged feature map $P$ at output stride 8 is used as the final feature map from Eff-SpineNet. We further apply separable convolutional layers before the pixel-level prediction layer. The number of layers and feature dimension for each layer are fixed to be the same as the subnets in object detection, shown in Table~\ref{table:scaling_map}.

\section{Experimental Results}
We present experimental results on object detection, image classification, and semantic segmentation to demonstrate the effectiveness and generality of the proposed Eff-SpineNet models.
For object detection, we evaluate Eff-SpineNet on COCO bounding box detection~\cite{coco}. We train all models on the COCO \texttt{train2017} split and report results on the COCO \texttt{val2017} split.
For image classification, we train Eff-SpineNet models on ImageNet ILSVRC-2012 and report Top-1 and Top-5 validation accuracy.
For semantic segmentation, we follow the common practice to train Eff-SpineNet on PASCAL VOC 2012 with augmented 10,582 training images and report mIOU on 1,449 \texttt{val} set images.
%train Eff-SpineNet models on PASCAL VOC 2012~\cite{pascalvoc2012} \texttt{train} set and report mIOU on PASCAL VOC 2012 \texttt{test} set. 

\subsection{Object Detection}
%%%%%%%%%%%%%%%%%%%%%%%%%%%%%%%%%%%%%%%%%%%%%
\setlength{\tabcolsep}{4pt}
\begin{table}[t]
\begin{center}
\caption{\textbf{One-stage object detection results on the COCO benchmark.} We compare using different backbones with RetinaNet on single model without test-time augmentation. FLOPs is represented by Multi-Adds}
\label{table:headings}
\scalebox{0.85}{
\begin{tabular}{c | c c | c c c c c c}
\hline
 model & \#FLOPs & \#Params & AP & AP$_{50}$ & AP$_{75}$ & AP$_{S}$ & AP$_{M}$ & AP$_{L}$\\[2pt]
 \hline
 
 \bf Eff-SpineNet-D0 & \bf 2.5B & \bf 3.6M & \bf 34.7 & \bf 53.1 & \bf 37.0 & \bf 15.2 & \bf 38.7 & \bf 52.8 \\
 EfficientNet-B0-FPN & 2.5B & 3.7M & 33.5 & 52.8 & 35.4 & 14.5 & 37.5 & 50.7 \\
 EfficientDet-D0~\cite{tan2020efficientdet} & 2.5B & 3.9M & 33.5 & - & - & - & - & - \\
 \hline
 \bf Eff-SpineNet-D1 & \bf 6.0B & \bf 5.2M & \bf 39.9 & \bf 59.6 & \bf 42.5 & \bf 43.5 & \bf 21.1 & \bf 57.5 \\
 EfficientNet-B1-FPN & 5.8B & 6.3M & 38.8 & 59.1 & 41.4 & 20.2 & 43.0 & 55.7 \\
 EfficientDet-D1~\cite{tan2020efficientdet} & 6.0B & 6.6M & 39.1 & - & - & - & - & - \\
 \hline
 \bf Eff-SpineNet-D2 & \bf 10.3B & \bf 6.2M & \bf 42.5 & \bf 62.0 & \bf 46.0 & \bf 24.5 & \bf 46.4 & \bf 57.6 \\
 EfficientNet-B2-FPN & 10.0B & 7.5M & 41.4 & 62.3 & 44.1 & 24.4 & 45.4 & 56.8 \\
 EfficientDet-D2~\cite{tan2020efficientdet} & 11.0B & 8.1M & 42.5 & - & - & - & - & - \\
 ResNet-50-FPN~\cite{Du2020SpineNet} & 96.8B & 34.0M & 42.3 & 61.9 & 45.9 & 23.9 & 46.1 & 58.5 \\
 SpineNet-49~\cite{Du2020SpineNet} & 85.4B & 28.5M & 44.3 & 63.8 & 47.6 & 25.9 & 47.7 & 61.1 \\[2pt]
\hline
\end{tabular}
}
\end{center}
\end{table}
\setlength{\tabcolsep}{1.4pt}
%%%%%%%%%%%%%%%%%%%%%%%%%%%%%%%%%%%%%%%%%%%%%

% %%%%%%%%%%%%%%%%%%%%%%%%%%%%%%%%%%%%%%%%%%%%%
% \setlength{\tabcolsep}{4pt}
% \begin{table}[t]
% \begin{center}
% \caption{Ablation studies on advanced training features for Eff-SpineNet-D2 and ResNet50-FPN. \textbf{SD:} stochastic depth; \textbf{MSTRAIN:} large-scale multi-scale training with a longer training schedule. If disabled, we use scale jittering [0.8, 1.2] to train for 72 epochs; \textbf{SE:} squeeze and excitation}
% \label{table:training_ablation}
% \scalebox{0.85}{
% \begin{tabular}{c | c c c c c }
% \hline
%  model & original &-SD & -swish & -MSTRAIN & -SE\\
%  \hline
%  Eff-SpineNet-D2 & 42.5 & 42.1 \textcolor{red}{(-0.4)} & 40.1 \textcolor{red}{(-2.0)} & 32.6 \textcolor{red}{(-7.5)} & 32.2 \textcolor{red}{(-0.4)} \\
%  ResNet-50-FPN & 42.3 & 40.7 \textcolor{red}{(-1.6)} & 40.4 \textcolor{red}{(-0.3)} & 37.0 \textcolor{red}{(-3.4)} & - \\
% \hline
% \end{tabular}
% }
% \end{center}
% \end{table}
% \setlength{\tabcolsep}{1.4pt}
% %%%%%%%%%%%%%%%%%%%%%%%%%%%%%%%%%%%%%%%%%%%%%

%%%%%%%%%%%%%%%%%%%%%%%%%%%%%%%%%%%%%%%%%%%%%
\setlength{\tabcolsep}{4pt}
\begin{table}[t]
\begin{center}
\caption{Ablation studies on advanced training strategies for Eff-SpineNet-D2 and ResNet50-FPN. We begins with 72 epochs training steps and multi-scale training [0.8, 1.2] as the baseline. \textbf{SE:} squeeze and excitation; \textbf{ms train:} large-scale multi-scale [0.1, 2.0] and extened training steps that attain the best performance (650 epochs for Eff-SpineNet-D2 and 250 epochs for ResNet-50-FPN); \textbf{Swish:} Swish activation that replaces ReLU; \textbf{SD:} stochastic depth}
\label{table:training_ablation}
\scalebox{0.85}{
\begin{tabular}{c | c c c c c }
\hline
 model & baseline & +SE & +ms train & +Swish & +SD  \\
 
 \hline
 Eff-SpineNet-D2 & 32.2& 32.6\textcolor{blue}{(+0.4)}
  &  40.1\textcolor{blue}{(+7.4)} & 42.1 \textcolor{blue}{(+2.0)} & 42.5\textcolor{blue}{(+0.4)} \\
 ResNet-50-FPN & 37.0 & N/A & 40.4 \textcolor{blue}{(+3.4)} & 40.7 \textcolor{blue}{(+0.3)} &  42.3\textcolor{blue}{(+1.6)}\\
\hline
\end{tabular}
}
\end{center}
\end{table}
\setlength{\tabcolsep}{1.4pt}
%%%%%%%%%%%%%%%%%%%%%%%%%%%%%%%%%%%%%%%%%%%%%

%%%%%%%%%%%%%%%%%%%%%%%%%%%%%%%%%%%%%%%%%%%%%
\setlength{\tabcolsep}{4pt}
\begin{table}[t]
\begin{center}
\caption{Impact of longer training schedule using advanced training strategies when training a model from scratch}
\label{table:training_epochs}
\scalebox{0.85}{
\begin{tabular}{c | c c c c c c}
\hline
 model & 72 epoch & 200 epoch & 350 epoch & 500 epoch & 650 epoch \\
 \hline
 Eff-SpineNet-D2 & 34.8 & 40.0 \textcolor{blue}{(+5.2)} & 41.4 \textcolor{blue}{(+1.4)} & 42.1 \textcolor{blue}{(+0.7)} & 42.5 \textcolor{blue}{(+0.4)} \\
\hline
\end{tabular}
}
\end{center}
\end{table}
\setlength{\tabcolsep}{1.4pt}
%%%%%%%%%%%%%%%%%%%%%%%%%%%%%%%%%%%%%%%%%%%%%

%%%%%%%%%%%%%%%%%%%%%%%%%%%%%%%%%%%%%%%%%%%%%
\setlength{\tabcolsep}{4pt}
\begin{table}
\begin{center}
\caption{\textbf{Two-stage object detection results on COCO.} We compare using different backbones with Mask R-CNN on single model}
\label{table:mask_rcnn}
\scalebox{0.85}{
\begin{tabular}{c | c c | c c c c c c}
\hline
 model & \#FLOPs & \#Params & AP & AP$_{50}$ & AP$_{75}$ & AP$^{\text{mask}}$ & AP$_{50}^{\text{mask}}$ & AP$_{75}^{\text{mask}}$\\
 \hline
 Eff-SpineNet-D0 & 4.7B & 4.6M & 35.0 & 54.0 & 37.3 & 30.5 & 50.2 & 32.2 \\
 \hline
 Eff-SpineNet-D1 & 9.2B & 6.4M & 40.7 & 60.9 & 44.1 & 35.0 & 56.9 & 36.8 \\
 \hline
 Eff-SpineNet-D2 & 16.0B & 9.2M & 42.9 & 63.5 & 46.5 & 37.3 & 60.2 & 39.1 \\
\hline
\end{tabular}
}
\end{center}
\end{table}
\setlength{\tabcolsep}{1.4pt}
%%%%%%%%%%%%%%%%%%%%%%%%%%%%%%%%%%%%%%%%%%%%%

%%%%%%%%%%%%%%%%%%%%%%%%%%%%%%%%%%%%%%%%%%%%%
\setlength{\tabcolsep}{4pt}
\begin{table}[t]
\begin{center}
\caption{\textbf{Mobile-size object detection results on COCO.} Eff-SpineNet models achieve the new state-of-the-art FLOPs \vs AP trade-off curve}
\label{table:ondevice}
\scalebox{0.85}{
\begin{tabular}{c | c c | c c c c}
\hline
 backbone model & \#FLOPs & \#Params & AP & AP$_{S}$ & AP$_{M}$ & AP$_{L}$\\
 \hline
 \bf Eff-SpineNet-M0 & \bf 0.15B & \bf 0.67M & \bf 17.3 & \bf 2.2 & \bf 16.4 & \bf 33.0 \\
 MobileNetV3-Small-SSDLite~\cite{mobilenetv3} & 0.16B & 1.77M & 16.0 & - & - & - \\
 \hline
 \bf Eff-SpineNet-M1 & \bf 0.51B & \bf 0.99M & \bf 25.0 & \bf 7.4 & \bf 27.3 & 42.0\\
 MobileNetV3-SSD~\cite{mobilenetv3} & 0.51B & 3.22M & 22.0 & - & - & - \\
 MobileNetV2 + MnasFPN & 0.53B & 1.29M & 23.8 & - & - & - \\
  MnasNet-A1-SSD~\cite{mnasnet}  & 0.8B & 4.9M & 23.0 & 3.8 & 21.7 & 42.0 \\
 \hline
 \bf Eff-SpineNet-M2 & \bf 0.97B & \bf 2.36M & \bf 29.2 & \bf 9.7 & \bf 32.7 & \bf 48.0 \\
 MobileNetV2-NAS-FPN~\cite{nasfpn}  & 0.98B & 2.62M & 25.7 & - & - & - \\
 MobileNetV2-FPN~\cite{mobilenetv2} & 1.01B & 2.20M & 24.3 & - & - & - \\
\hline
\end{tabular}
}
\end{center}
\end{table}
\setlength{\tabcolsep}{1.4pt}
%%%%%%%%%%%%%%%%%%%%%%%%%%%%%%%%%%%%%%%%%%%%%

%%%%%%%%%%%%%%%%%%%%%%%%%%%%%%%%%%%%%%%%%%%%%
\begin{figure}[t]
\centering
%\begin{minipage}{.48\textwidth}
  \centering
  \includegraphics[height=1.6in]{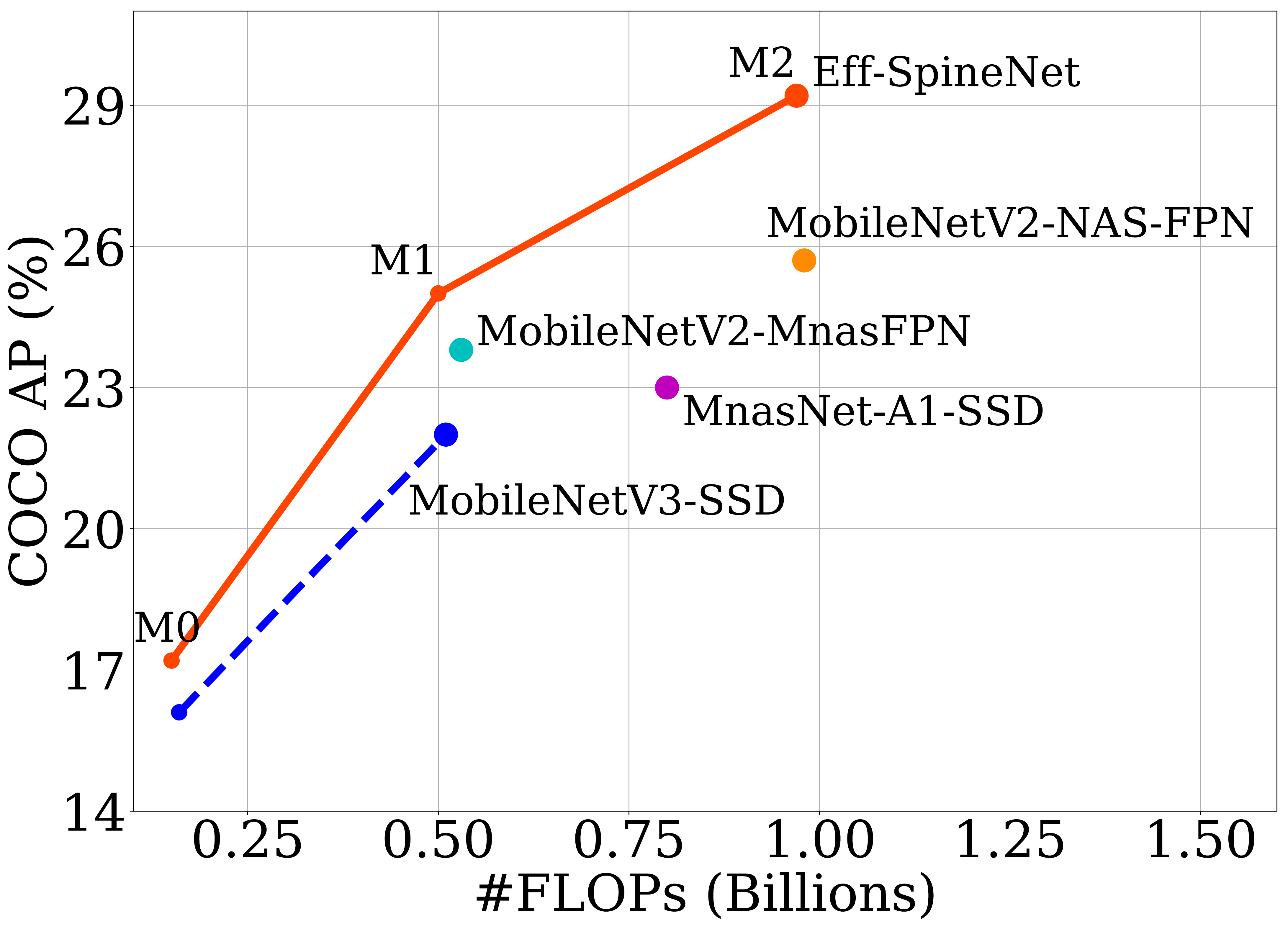}
  \captionof{figure}{A comparison of Eff-SpineNet and other state-of-the-art detectors on mobile-size object detection. Eff-SpineNet models outperform the other detectors at various scales}
  \label{fig:flops_ap_params}
%\end{minipage}%
%\hspace{10}
%\begin{minipage}{.48\textwidth}
%  \centering
%  \includegraphics[height=1.6in]{imgs/spv2_cls_params.pdf}
%  \captionof{figure}{A comparison of Params \vs Accuracy of Eff-SpineNet and other image %classification models. Detailed comparison between Eff-SpineNet and EfficientNet can be %found in Table~\ref{table:classification}}
%  \label{fig:acc_params}
%\end{minipage}
\end{figure}
%%%%%%%%%%%%%%%%%%%%%%%%%%%%%%%%%%%%%%%%%%%%%

\subsubsection{Experimental Settings}
\label{sec:detection_settings}

\paragraph{\bf Training details:} We generally follow the training protocol in~\cite{Du2020SpineNet,tan2020efficientdet} to train all models for the proposed method, EfficientNet-FPN, and SpineNet on COCO \texttt{train2017} from scratch. We train on Cloud TPU v3 devices using standard stochastic gradient descent (SGD) with 4e-5 weight decay and 0.9 momentum. We apply batch size 256 and stepwise learning rate with 0.28 initial learning rate that decays to $0.1\times$ and $0.01\times$ at the last 30 and 10 epochs. All models are trained for 650 epochs, which we observe the model starts to overfit and hurt performance after 650 epochs. We apply synchronized batch normalization with $0.99$ momentum, swish activation~\cite{ramach2017searching}, and stochastic depth~\cite{dropconnect}. To pre-process training data, we resize the long side of an image to the target image size described in Table~\ref{table:scaling_map} then pad the short side with zeros to make it a square image. Horizontal flipping and multiscale augmentation [0.1, 2.0] are implemented during training.

\paragraph{\bf Search details:} We design our search space $\{\beta_1,\beta_2,...,\beta_N\}$ as $\{1,5,10,15,20\}$ in this work to cover a wide range of possible resource distributions with a manageable search space size. We follow~\cite{tan2019efficientnet,Du2020SpineNet} to implement the reinforcement learning based search method~\cite{nas}. In brief, we reserve $7392$ images from COCO \texttt{train2017} as the validation set for searching and use other images for training. Sampled models at the D0 scale are used for proxy task training with the same training settings described above. AP on the reserved set of proxy tasks trained for 4.5k iterations is collected as rewards. The best architecture is collected after 5k architectures have been sampled.

\subsubsection{Object Detection Results}
\label{sec:detection_results}

\paragraph{\bf RetinaNet:} 
Our main results are presented on the COCO bounding box detection task with RetinaNet. Compared to our architecture-wise baseline EfficientNet-FPN models, our models consistently achieve 1-2\% AP gain from scale D0 to D2 while using less computations. The FLOPs \vs AP curve and the Params \vs AP curve among Eff-SpineNet and other state-of-the-art one-stage object detectors are shown in Figure~\ref{fig:flops_ap_curve} and Figure~\ref{fig:flops_ap_params} respectively.

%%%%%%%%%%%%%%%%%%%%%%%%%%%%%%%%%%%%%%%%%%%%%
%\begin{figure}
%\centering
%\includegraphics[height=1.6in]{}
%\caption{Params \vs AP curves of Eff-SpineNet, %EfficientNet-FPN, and EfficientDet. Eff-SpineNet models %outperform the other two with large margin at all %scales}
%\label{fig:flops_ap_params}
%\end{figure}
%%%%%%%%%%%%%%%%%%%%%%%%%%%%%%%%%%%%%%%%%%%%%

\paragraph{\bf Mask R-CNN:}
We evaluate Eff-SpineNet models with Mask R-CNN on the COCO bounding box detection and instance-level segmentation task.
The results of Eff-SpineNet D0, D1, and D2 models are shown in Table~\ref{table:mask_rcnn}.

\subsubsection{Mobile-size Object Detection Results\newline\newline}
The results of Eff-SpineNet-M0/1/2 models are presented in Table~\ref{table:ondevice} and the FLOPs \vs AP curve is plotted in Figure~\ref{fig:flops_ap_curve}. Eff-SpineNet models are able to consistent use less resources while surpassing all other state-of-the-art mobile-size object detectors by large margin. In particular, our Eff-SpineNet-M2 achieves 29.2\% AP with 0.97B FLOPs, attaining the new state-of-the-art for mobile-size object detection.

\subsubsection{Ablation Studies}
\paragraph{\bf Ablation studies on advanced training strategies:} 
We conduct detailed ablation studies on the advanced training features used in this paper and~\cite{tan2019efficientnet,Du2020SpineNet}. Starting from the final Eff-SpineNet-D2 model, we gradually remove one feature at a time: 1) removing stochastic depth in model training leads to 0.4 AP drop; 2) replacing swish activation with ReLU leads to 2.0 AP drop; 3) using less aggressive multi-scale training strategy with 72 training epochs leads to 7.5 AP drop; 4) removing squeeze and excitation~\cite{squeezeexcitation} layers from all MBConv blocks leads to 0.4 AP drop. We further perform the ablation studies to ResNet-50-FPN and the results are shown in Table~\ref{table:training_ablation}.

\paragraph{\bf Impact of longer training schedule:} 
We conduct experiments by adopting different training epochs for Eff-SpineNet-D2. We train all models from scratch on COCO 2017train and report AP on COCO 2017val. The results are presented in Table~\ref{table:training_epochs}. We show that prolonging the training epochs from 72 to 650 gradually improves the performance of Eff-SpineNet-D2 by 7.7\% AP. Except training schedule, the other training strategies are the same as Section~\ref{sec:detection_settings}.

\paragraph{\bf Learning Resource Distribution:} From the final architecture discovered by NAS shown in Table~\ref{table:architecture}, we observe that parameters in low-level $L_2$ blocks and high-level \{$L_6$, $L_7$\} block, are reallocated to middle-level \{$L_3$, $L_4$, $L_5$\} blocks. Since the high-level blocks are low in resolution, by doing so, the number of parameters in the network is significantly reduced from 4.4M to 3.6M while the total FLOPs remains roughly the same. Learning resource distribution also brings a 0.8\% AP gain. The performance improvements from SP-EfficientNet-B0 to Eff-SpineNet-D0 is shown in Table~\ref{table:ablation_dist}.

%%%%%%%%%%%%%%%%%%%%%%%%%%%%%%%%%%%%%%%%%%%%%
\setlength{\tabcolsep}{4pt}
\begin{table}[t]
\begin{center}
\caption{Improvement from learning a better resource distribution. All models are evaluated on COCO \texttt{2017val}}
\label{table:ablation_dist}
\scalebox{0.85}{
\begin{tabular}{c | c c | c c c c c c}
\hline
 model & \#FLOPs & \#Params & AP & AP$_{50}$ & AP$_{75}$ & AP$_{S}$ & AP$_{M}$ & AP$_{L}$\\
 \hline
 SP-EfficientNet-B0 & 2.4B & 4.4M & 33.0 & 50.3 & 34.7 & 13.0 & 38.4 & 51.7 \\
 Eff-SpineNet-D0 & 2.5B & 3.6M & 33.8 & 51.3 & 35.8 & 13.6 & 39.3 & 52.4\\
 \hline
% \hline
% MobileNetV3-Small & 0.21B & 2.49M & 16.0 \\[2pt]
% MobileNetV3-Small (modified) & 0.16B & 1.77M & 16.1 \\[2pt]
%\hline
\end{tabular}
}
\end{center}
\end{table}
\setlength{\tabcolsep}{1.4pt}
%%%%%%%%%%%%%%%%%%%%%%%%%%%%%%%%%%%%%%%%%%%%%

%%%%%%%%%%%%%%%%%%%%%%%%%%%%%%%%%%%%%%%%%%%%%
\setlength{\tabcolsep}{4pt}
\begin{table}[t]
\begin{center}
\caption{An ablation study of the two architecture improvements in Eff-SpineNet}
\label{table:ablation_2tech}
\scalebox{0.85}{
\begin{tabular}{c | c c | c c | c}
\hline
 model & weighted fusion & space-to-depth  & \#FLOPs & \#Params & AP \\[2pt]
 \hline
 Eff-SpineNet-D0 & \checkmark & \checkmark & 2.5B & 3.6M & 33.8\\
 model 1 & \checkmark & - & 2.4B & 3.3M & 33.1\\
 model 2 & - & - & 2.4B & 3.3M & 32.8\\
\hline
\end{tabular}
}
\end{center}
\end{table}
\setlength{\tabcolsep}{1.4pt}
%%%%%%%%%%%%%%%%%%%%%%%%%%%%%%%%%%%%%%%%%%%%%

\paragraph{\bf Architecture Improvements:} We conduct ablation studies for the two techniques, resampling method based on the space-to-depth operation and weighted block fusion, introduced to SpineNet's scale-permuted architecture with Eff-SpineNet-D0. As shown in Table~\ref{table:ablation_2tech}, the performance drops by 0.7\% AP if we remove the new resampling method. The performance further drops by 0.3\% AP if we remove weighted block fusion. 

%%%%%%%%%%%%%%%%%%%%%%%%%%%%%%%%%%%%%%%%%%%%%
\begin{figure}[t]
\centering
\begin{subfigure}[t]{0.5\textwidth}
        \centering
        \includegraphics[height=1.65in]{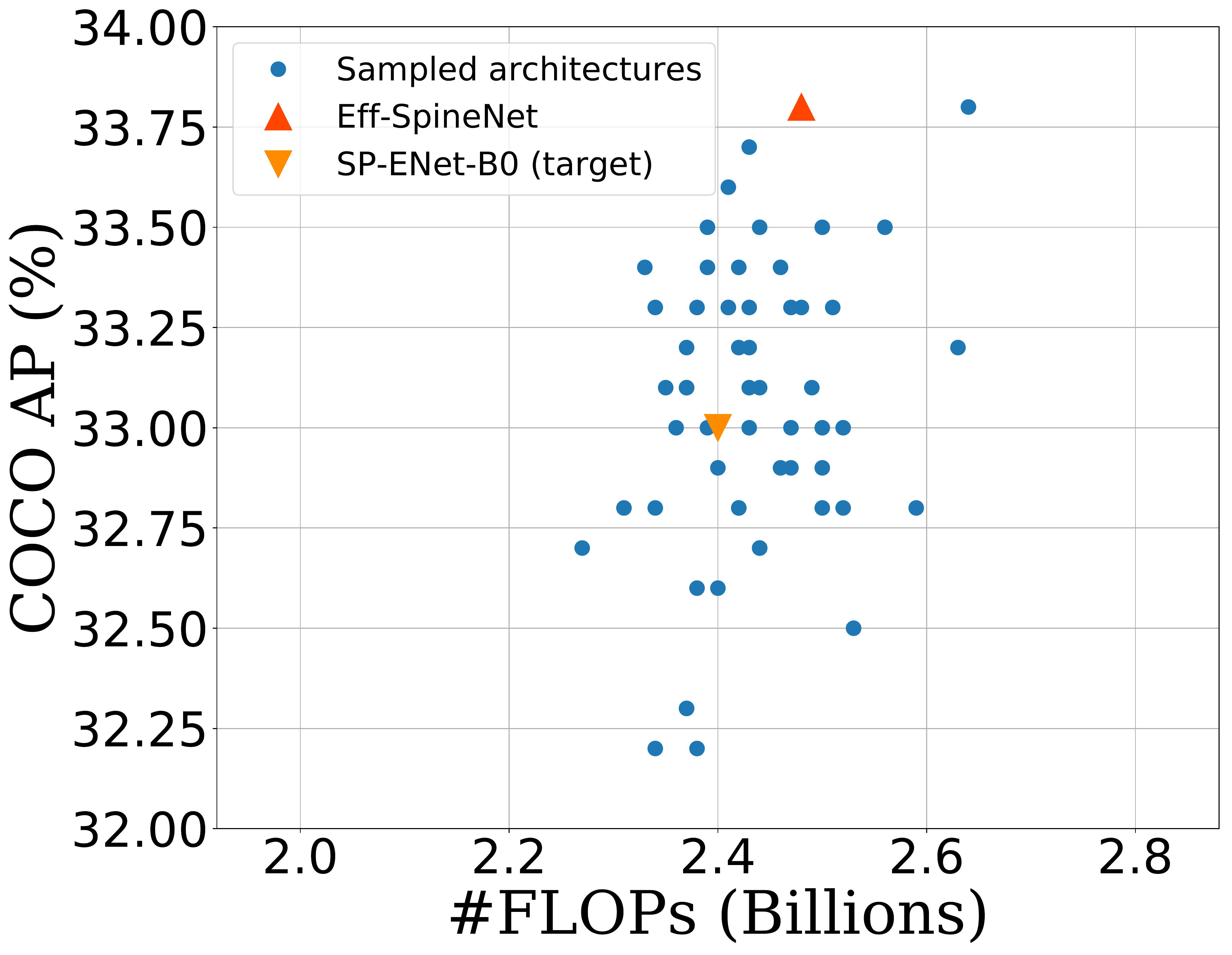}
    \end{subfigure}%
\begin{subfigure}[t]{0.5\textwidth}
        \centering
        \includegraphics[height=1.65in]{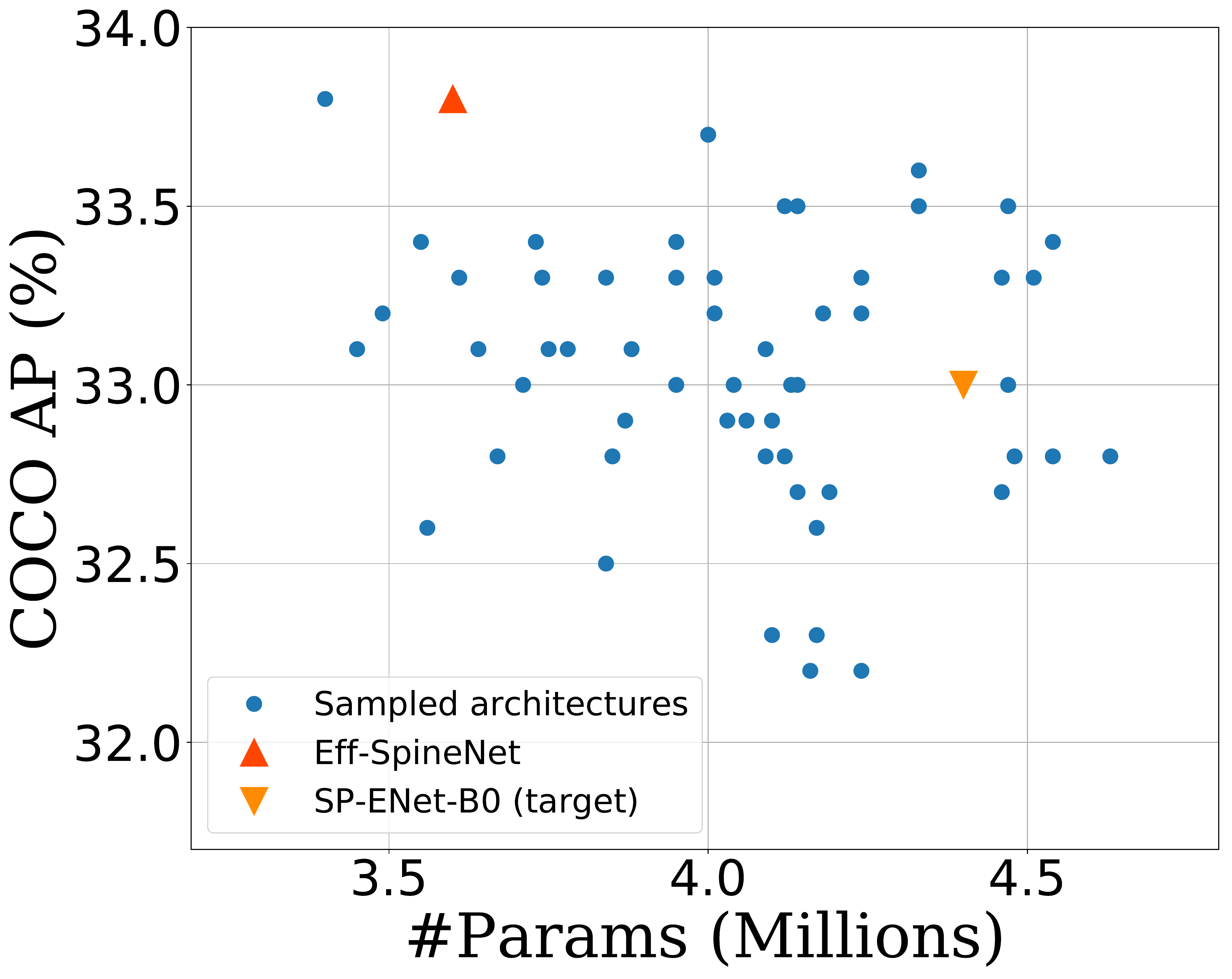}
    \end{subfigure}%
\caption{FLOPs \vs AP (left) and Params \vs AP (right) plots for architectures sampled throughout the searching phase. The $x$-axes are plotted within a $\pm20\%$ range to the centers}
\label{fig:search_plot}
\end{figure}
%%%%%%%%%%%%%%%%%%%%%%%%%%%%%%%%%%%%%%%%%%%%%

\subsubsection{A study of the proposed search algorithm\newline\newline}
We visualize some of the randomly sampled architectures in the search phase. The FLOPs \vs AP plot and the Params \vs AP plot are presented in Figure~\ref{fig:search_plot}. From the FLOPs \vs AP plot, we can observe that the FLOPs of the sampled architectures fall into a range of $\pm10\%$ of our target FLOPs because of the proposed search algorithm. We can also observe from the Params \vs AP plot that the number of parameters in sampled architectures are reduced in most cases.

\subsection{Image Classification}
We conduct image classification experiments on ImageNet ILSVRC 2012~\cite{deng09imagenet,ilsvrc} with Eff-SpineNet, following the same training strategy used in EfficientNet~\cite{tan2019efficientnet}.
We scale the input size with respect to different model size by roughly following the compound scaling~\cite{tan2019efficientnet} and adjusting it to be the closest multiples of 16.

% First, we perform an ablation study to compare SP-EfficientNet with Eff-SpineNet and decide the feature dimension for the classification model.
% From results in Table~\ref{table:classification_ablation}, we observe that 1) the learned resource distribution is able to reduce the number of parameters while slightly increase the performance and 2) feature dimension higher than $256$ doesn't improve the model performance.
% Therefore, we use Eff-SpineNet with learned resource distribution and feature dimension of $256$ for all image classification experiments. 

%%%%%%%%%%%%%%%%%%%%%%%%%%%%%%%%%%%%%%%%%%%%%
%\setlength{\tabcolsep}{4pt}
%\begin{table}[t]
%\caption{Image classification ablation study on resource distribution and feature dimension. All models take the input resolution of $224\times224$}
%\label{table:classification_ablation}
%\begin{center}
%\scalebox{0.85}{
%\begin{tabular}{c | c | c c | c c }
%\hline
%model & feature dim & \#FLOPs & \#Params & Top-1 & Top-5 \\[2pt]
%\hline
%SP-EfficientNet-B0 & 256 & 0.36B & 4.23M & 74.9 & 92.0 \\
%\hline
%Eff-SpineNet-D0 & 128 & 0.38B & 3.38M & 74.8 & 92.1 \\
%\textbf{Eff-SpineNet-D0} & \textbf{256} & \textbf{0.38B} & \textbf{3.57M} & \textbf{75.3} & %\textbf{92.4} \\
%Eff-SpineNet-D0 & 512 & 0.40B & 3.94M & 75.3 & 92.4 \\
%Eff-SpineNet-D0 & 1024 & 0.44B & 4.69M & 75.4 & 92.4 \\
%\hline
%\end{tabular}
%}
%\end{center}
%\end{table}
%\setlength{\tabcolsep}{1.4pt}
%%%%%%%%%%%%%%%%%%%%%%%%%%%%%%%%%%%%%%%%%%%%%

%%%%%%%%%%%%%%%%%%%%%%%%%%%%%%%%%%%%%%%%%%%%%
\setlength{\tabcolsep}{4pt}
\begin{table}[t]
\caption{Comparison between Eff-SpineNet and EfficientNet on ImageNet classification}
\label{table:classification}
\begin{center}
\scalebox{0.85}{
\begin{tabular}{c | c | c | c c | c c }
\hline
 model & input resolution & feature dim & \#FLOPs & \#Params & Top-1 & Top-5 \\[2pt]
 \hline
 \textbf{Eff-SpineNet-D0} & \textbf{$224\times224$} & \textbf{256} & \textbf{0.38B} & \textbf{3.57M} & \textbf{75.3} & \textbf{92.4} \\
 EfficientNet-B0 & $224\times224$ & 1280 & 0.39B & 5.30M & 77.3 & 93.5 \\
 \hline
 \textbf{Eff-SpineNet-D1} & \textbf{$240\times240$} & \textbf{256} & \textbf{0.70B} & \textbf{4.97M} & \textbf{77.7} & \textbf{93.6} \\
 EfficientNet-B1 & $240\times240$ & 1280 & 0.70B & 7.80M & 79.2 & 94.5 \\
 \hline
 \textbf{Eff-SpineNet-D2} & \textbf{$256\times256$} & \textbf{256} & \textbf{0.89B} & \textbf{5.83M} & \textbf{78.5} & \textbf{94.2} \\
 EfficientNet-B2 & $260\times260$ & 1280 & 1.00B & 9.20M & 80.3 & 95.0 \\
\hline
\end{tabular}
}
\end{center}
\end{table}
\setlength{\tabcolsep}{1.4pt}
%%%%%%%%%%%%%%%%%%%%%%%%%%%%%%%%%%%%%%%%%%%%%

We compare Eff-SpineNet with EfficientNet in all aspects in Table~\ref{table:classification}.
At the same FLOPs, Eff-SpineNet is able to save around 35\% parameters at the cost of 1.5-2\% drop in top-1 accuracy.
We hypothesize this is likely due to the fact that higher level features ($P_6$ and $P_7$) do not contain enough spatial resolution for small input size. For $224\times224$ input size, the spatial resolution of $P_6$ and $P_7$ is only $4\times4$ and $2\times2$ respectively.
We will explore how to construct better scale-permuted models for image classification in the future.

\subsection{Semantic Segmentation}

%%%%%%%%%%%%%%%%%%%%%%%%%%%%%%%%%%%%%%%%%%%%%
\setlength{\tabcolsep}{4pt}
\begin{table}[t]
\begin{center}
\caption{Semantic segmentation result comparisons of Eff-SpineNet and other popular semantic segmentation networks on the PASCAL VOC 2012 \texttt{val} set}
\label{table:seg}
\scalebox{0.85}{
\begin{tabular}{c| c c | c  c c }
\hline
 model & \begin{tabular}[x]{@{}c@{}}ImageNet\\pre-train\end{tabular} & \begin{tabular}[x]{@{}c@{}}COCO\\pre-train\end{tabular} & \begin{tabular}[x]{@{}c@{}}output\\stride\end{tabular} & \#FLOPs & mIOU \\[2pt]
 \hline
 MobileNetv2 + DeepLabv3 & - & \checkmark & 16 & 2.8B & 75.3 \\[2pt]
 ResNet-101 + DeepLabv3 &  \checkmark & \checkmark & 8 & 81.0B & 80.5 \\[2pt]
 \hline
 \bf Eff-SpineNet-D0 & - & \checkmark & 8 & \bf 2.1B & \bf 76.0 \\[2pt]
 \bf Eff-SpineNet-D2 & - & \checkmark & 8 & \bf 3.8B & \bf 78.0 \\[2pt]

\hline
\end{tabular}
}
\end{center}
\end{table}
\setlength{\tabcolsep}{1.4pt}
%%%%%%%%%%%%%%%%%%%%%%%%%%%%%%%%%%%%%%%%%%%%%

We present experimental results of employing Eff-SpineNet as backbones for semantic segmentation. We conduct the experiments with evaluation metric mIOU on PASCAL VOC 2012~\cite{pascalvoc2012} with extra annotated images from~\cite{extra_seg_data}. For training implementations, we generally follow the settings in Section~\ref{sec:detection_settings}. In brief, we fine-tune all models from the COCO bounding box detection pre-trained models for 10k iterations with batch size 256 with cosine learning rate. We set the initial learning to 0.05 and a linear learning rate warmup is applied for the first 500 iterations. We fix the input crop size to $512\times 512$ for all Eff-SpineNet models without strictly following the scaling rule described in Table~\ref{table:scaling_map}. 

Our results on PASCAL VOC 2012 \texttt{val} set are presented in Table~\ref{table:seg}. Eff-SpineNet-D0 slightly outperforms MobileNetv2 with DeepLabv3~\cite{mobilenetv2,Chen2017deeplabv3} by 0.7 mIOU while using 25\% less FLOPs. Our D2 model is able to attain comparable mIOU with other popular semantic segmentation networks, such as ResNet101 with DeepLabv3~\cite{Chen2017deeplabv3}, at the same output stride while using 95\% less FLOPs.

\section{Conclusion}
In this paper, we propose to decompose model design into architecture topology, operation, and resource distribution. We show that simply combining scale-permuted architecture topology and efficient operations achieves new state-of-the-art in object detection, showing the benefits of efficient operation and scale-permuted architecture are complementary. The model can be further improved by learning the resource distribution over the entire network. The resulting Eff-SpineNet is a versatile backbone model that can be also applied to image classification and semantic segmentation tasks, attaining competitive performance, proving Eff-SpineNet is a versatile backbone model that can be easily applied to many tasks without extra architecture design.

\clearpage
\bibliographystyle{splncs04}
\bibliography{arxiv}
\end{document}

% \linenumbers
\pagestyle{headings}
\mainmatter
\def\ECCVSubNumber{4374}  % Insert your submission number here

% \title{SpineNetv2: Building SpineNet with Efficient Building s (placeholder)} % Replace with your title
\title{Efficient Scale-Permuted Backbone with Learned Resource Distribution}

%\begin{comment}
\titlerunning{ECCV-20 submission ID \ECCVSubNumber} 
\authorrunning{ECCV-20 submission ID \ECCVSubNumber} 
\author{Anonymous ECCV submission}
\institute{Paper ID \ECCVSubNumber}
%\end{comment}
%******************

\begin{comment}
\titlerunning{Abbreviated paper title}
\author{First Author\inst{1}\orcidID{0000-1111-2222-3333} \and
Second Author\inst{2,3}\orcidID{1111-2222-3333-4444} \and
Third Author\inst{3}\orcidID{2222--3333-4444-5555}}
%
\authorrunning{F. Author et al.}
\institute{Princeton University, Princeton NJ 08544, USA \and
Springer Heidelberg, Tiergartenstr. 17, 69121 Heidelberg, Germany
\email{lncs@springer.com}\\
\url{http://www.springer.com/gp/computer-science/lncs} \and
ABC Institute, Rupert-Karls-University Heidelberg, Heidelberg, Germany\\
\email{\{abc,lncs\}@uni-heidelberg.de}}
\end{comment}
%******************
\maketitle

\appendix
\section{Effect of using different training iterations}
In this section, we conduct experiments by adopting different training iterations for SpineNetv2-D1 and its baseline EfficientNet-B1-FPN. We train all models from scratch on COCO \texttt{2017train} and report AP on COCO \texttt{2017val}. The results are presented in Table~\ref{table:iterations}. We show that prolonging the training iterations from 50k to 350k improves the performance of SpineNetv2-D1 by 4.1\% AP and EfficientNet-B1-FPN by 3.6\% AP. EfficientNet-B1-FPN converges around iteration 200k and won't benefit from a longer training schedule. SpineNetv2 is further improved by 0.9\% AP by increasing the training iteration from 200k to 350k. 

%%%%%%%%%%%%%%%%%%%%%%%%%%%%%%%%%%%%%%%%%%%%%
\setlength{\tabcolsep}{4pt}
\begin{table}
\begin{center}
\caption{Effect of using different training iterations for SpineNetv2 and EfficientNet-B1-FPN. All models are train on COCO \texttt{2017train} and evaluate on COCO \texttt{2017val}}
\label{table:iterations}
\vspace{3}
\scalebox{0.85}{
\begin{tabular}{c | c c c c c}
\hline
 model & iter. 50k & iter. 100k & iter. 200k & iter. 280k & iter. 350k\\
 \hline
 SpineNetv2-D1 & 35.3 & 37.5 & 38.5 & 39.2 & 39.4\\
 EfficientNet-B1-FPN & 33.7 & 36.7 & 37.3 & 37.1 & 37.1 \\
 \hline
\end{tabular}
}
\end{center}
\vspace{-20pt}
\end{table}
\setlength{\tabcolsep}{1.4pt}
%%%%%%%%%%%%%%%%%%%%%%%%%%%%%%%%%%%%%%%%%%%%%

%\bibliographystyle{splncs04}
\bibliography{egbib}